\documentclass{INTERSPEECH2023}

\interspeechcameraready 

\usepackage{times}
\usepackage{latexsym}
\usepackage[T1]{fontenc}
\usepackage[utf8]{inputenc}
\usepackage{microtype}
\usepackage{url}
\usepackage{hyperref}
\usepackage{amsmath,amsfonts}
\usepackage{algorithm}
\usepackage{algorithmic}
\usepackage{arydshln}
\usepackage{multirow,subfigure,booktabs}
\usepackage{caption}
\usepackage{amssymb}
\usepackage{graphicx}

\title{Knowledge-Retrieval Task-Oriented Dialog Systems with Semi-Supervision}
\name{Yucheng Cai$^{1,3}$, Hong Liu$^{1,3}$, Zhijian Ou$^{*,1,3}$\thanks{$^{*}$ Corresponding author: Zhijian Ou. The code is released at \url{https://github.com/thu-spmi/JSA-KRTOD}
}, Yi Huang$^{2,3}$, Junlan Feng$^{2,3}$
}
\address{
$^{1}$Speech Processing and Machine Intelligence (SPMI) Lab, Tsinghua University, Beijing, China \\
  $^{2}$China Mobile Research Institute, Beijing, China \\
  $^{3}$Tsinghua University-China Mobile Communications Group Co., Ltd. Joint Institute, Beijing, China
}
\email{ \{cyc22,liuhong21\}@mails.tsinghua.edu.cn,ozj@tsinghua.edu.cn, \{huangyi,fengjunlan\}@chinamobile.com}
\date{October 2022}
\newcommand{\modelname}{JSA-KRTOD}

\newcommand{\comparemodel}{PL}

\begin{document}

\maketitle
\vspace{-0.5em}
\begin{abstract}
\vspace{-0.5em}
Most existing task-oriented dialog (TOD) systems track dialog states in terms of slots and values and use them to query a database to get relevant knowledge to generate responses.
In real-life applications, user utterances are noisier, and thus it is more difficult to accurately track dialog states and correctly secure relevant knowledge.
Recently, a progress in question answering and document-grounded dialog systems is retrieval-augmented methods with a knowledge retriever.
Inspired by such progress, we propose a retrieval-based method to enhance knowledge selection in TOD systems, which significantly outperforms the traditional database query method for real-life dialogs. Further, we develop latent variable model based semi-supervised learning, which can work with the knowledge retriever to leverage both labeled and unlabeled dialog data. 
Joint Stochastic Approximation (JSA) algorithm is employed for semi-supervised model training, and the whole system is referred to as \modelname{}. Experiments are conducted on a real-life dataset from China Mobile Custom-Service, called MobileCS, and show that \modelname{} achieves superior performances in both labeled-only and semi-supervised settings.

\end{abstract}
\noindent\textbf{Index Terms}: task-oriented dialog system, knowledge retrieval, semi-supervised learning, joint stochastic approximation
\vspace{-0.5em}
\section{Introduction}

\begin{figure}[t]
\centering
 \subfigure[Traditional TOD system]{
\includegraphics[width=0.99\linewidth]{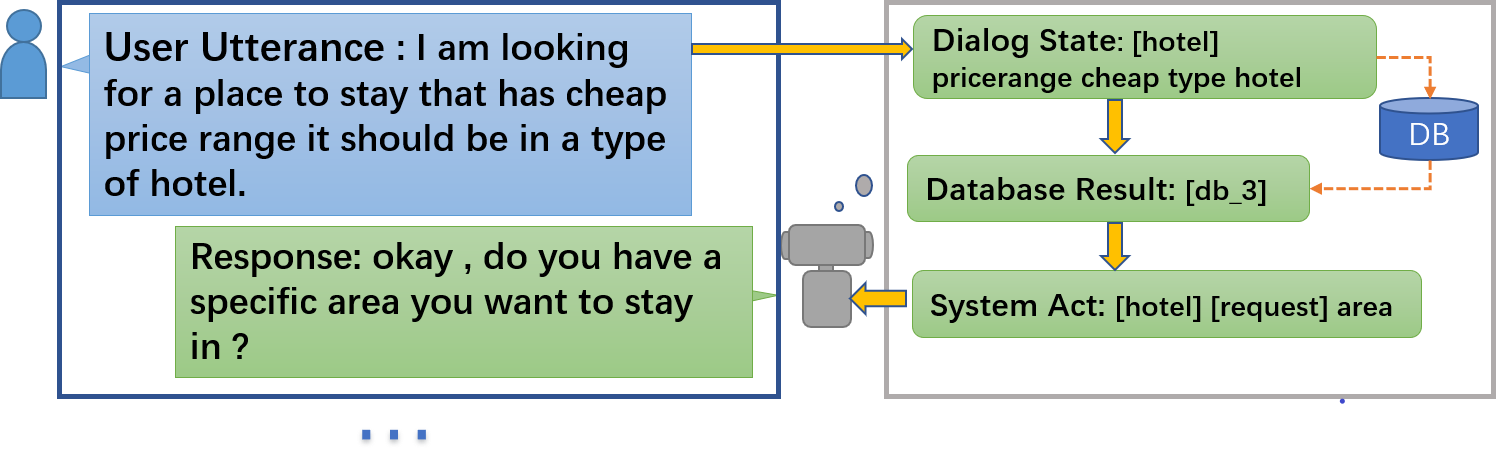}}
\vspace{-0.5em}
\subfigure[Our KRTOD system]{
\includegraphics[width=0.99\linewidth]{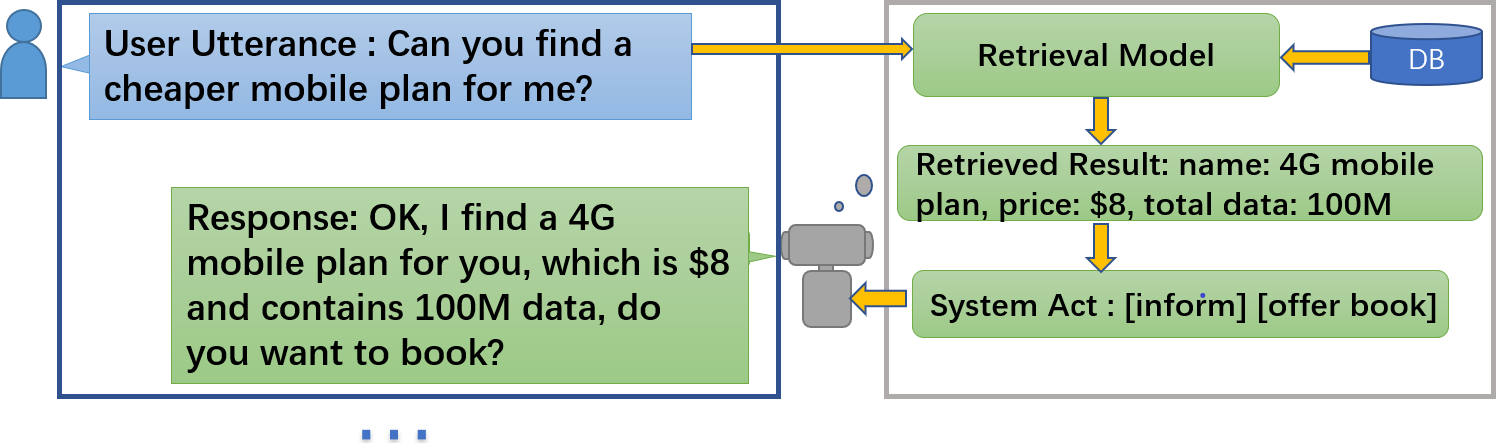}}
 	\vspace{-0.5em}
	\caption{Comparison between our KRTOD system and the traditional TOD system. KRTOD uses a neural network based retrieval model instead of the traditional rule-based database query, and can yield more informative and successful responses.}
 	\vspace{-1.0em}
	\label{fig:overview}
 \vspace{-1.0em}
\end{figure}

Task-oriented dialog (TOD) systems are designed to help users to achieve their goals through multiple turns of natural language interaction \cite{williams2016dialog,zhang2020task}.
The system needs to perform dialog state tracking (DST), query a task-related database (DB), decide actions and generate responses iteratively across turns.
The information flow in a task-oriented dialog is illustrated in Figure \ref{fig:overview}. 

Recent studies recast such information flow in TOD systems as conditional generation of tokens based on pretrained language models (PLMs) such as GPT2 \cite{radford2019gpt2} and T5 \cite{raffel2020t5}. Fine-tuning PLMs over annotated dialog datasets via supervised learning \cite{hosseini2020simple, peng2020etal, liu2022mga,lee2021improving} has shown promising results on Wizard-of-Oz TOD datasets such as MultiWOZ \cite{budzianowski2018large}. 
However, the Wizard-of-Oz dialog data are in fact simulated data.
In real-life applications such as custom services, user utterances are more casual and may contain noises from automatic speech recognition.
It is more difficult for the dialog system to perform well on DST, and the rule-based database query is not robust to erroneous dialog states. This brings performance degradation in real-life applications.

To address the problems mentioned above, we propose to introduce a knowledge retriever into TOD systems, and the new system is called \textbf{k}nowledge-\textbf{r}etrieval TOD system (KRTOD). It has been shown in recent studies \cite{lewis2020retrieval, kang2022knowledge, qu2021rocketqa} that knowledge retrievers, such as BM25 or dense passage retriever (DPR), can retrieve appropriate external knowledge based on a conversational context in question answering and knowledge grounded dialog. By introducing the retrievers, both the difficulty in correctly tracking the dialog state and the inflexibility of rule-based database query can be alleviated in KRTOD. First, through using the retriever to secure knowledge from the database, we can avoid tracking of dialog states, which originally are needed for database query. Annotations for DST may also be omitted, simplifying the procedure of building TOD systems. Second, using a retriever instead of rule-based database query makes the system more flexible in knowledge query. More complex situations, such as that illustrated in Figure \ref{fig:overview} can be handled through a retriever. 

Further, note that obtaining intermediate labels of the ground-truth knowledge is usually expensive and difficult for knowledge-grounded systems.
So for KRTOD systems, we further develop latent variable model based semi-supervised learning, which can work with the knowledge retriever to leverage both labeled and unlabeled dialog data. The Joint Stochastic Approximation (JSA) algorithm \cite{ou2020joint,cai2022advancing} is employed for semi-supervised model training.
The whole system is called \modelname{}. 
Intuitively, when the knowledge source is not available (i.e., over unlabeled data), the inference model in JSA learning help to infer the required knowledge from the system response. 

Experiments are conducted on a real-life human-human dialog dataset, called MobileCS (Mobile Customer-Service), released from the EMNLP 2022 SereTOD Challenge \cite{ou2022achallenge}. It consists of real-world dialog transcripts between real users and customer-service staffs from China Mobile, which are more
noisy and casual than prior Wizard-of-Oz data.

In summary, main contributions of this work are three folds:
\begin{itemize}
\item We propose to use a knowledge 
retriever to retrieve knowledge from the database instead of the traditional database query method in TOD system. The proposed KRTOD system is more suitable for real-life applications with nosier user utterances.

\item Further, we propose to use the JSA algorithm to perform semi-supervised learning for KRTOD systems. The resulting \modelname{} system can work with the knowledge retriever to effectively leverage both labeled and
unlabeled dialog data.

\item Extensive experiments are conducted on a real-life dataset, MobileCS, and show that \modelname{} achieves superior performances in both labeled-only and semi-supervised settings.
 \end{itemize}

\vspace{-0.5em}
\section{Related Work}
\vspace{-0.5em}
\subsection{TOD Systems}
\vspace{-0.5em}
Recent studies recast TOD systems as an end-to-end conditional generation problem based on pretrained language models (PLMs) such as GPT2 \cite{radford2019gpt2} and T5 \cite{raffel2020t5}. Those systems concatenate user utterances, dialog states, DB results, system acts and responses and train the model to generate them autoregressively \cite{hosseini2020simple, peng2020etal,liu2022mga, lee2021improving}, which achieves good performance on TOD datasets such as MultiWOZ \cite{budzianowski2018large}. Dialog state tracking \cite{heck2020trippy, lee2021dialogue} is crucial for the whole system, and the dialog state is used to query the database to get necessary information for appropriate response generation. However, previous studies have pointed out that those systems perform poorly in real-life applications \cite{liu2022information, zeng2022semi}, mainly because of poor results from database querying caused by inaccurate dialog states. The work in \cite{zeng2022semi} proposes to prepend the whole local knowledge base to the context, instead of doing database query. Their method achieves competent performance; However, it is not scalable with large knowledge bases. 
Our method makes the contribution that we introduce a knowledge retriever into the system to improve the quality of knowledge selection.
\vspace{-0.5em}
\subsection{Knowledge Retriever for Conditional Generation}
\vspace{-0.5em}
Recent researches such as RAG \cite{lewis2020retrieval} and REALM \cite{guu2020realm} have introduced knowledge retrieval models into conditional generation, which greatly improves the quality of generated responses in knowledge-intensive tasks such as open-domain question answering and knowledge-grounded 
 dialog systems. Those works use BM25 or neural network based retrievers such as DPR \cite{karpukhin-etal-2020-dense} to retrieve relevant knowledge given context and generate the answer using the retrieved knowledge pieces. 
There are several recent studies that improve over the original retrieval-augmented generation systems. Poly-encoder \cite{humeau2020poly} proposes to use a poly encoder instead of the original dual encoder to improve the retrieval accuracy.
Fusion-in-Decoder \cite{izacard2021leveraging} proposes to process retrieved passages independently in the encoder, but jointly fused them in the decoder, which improves the generation quality. 
SeekeR \cite{shuster2022language} and BlenderBot3 \cite{shuster2022blenderbot} recently propose to train a model to generate better query for retrieving knowledge.
However, none of these previous works apply a retrieval model to TOD systems.
Our work makes the contribution that we introduce a knowledge retriever into TOD system to substitute the original dialog state tracking and database query modules, which improves the accuracy of knowledge selection.
\vspace{-0.5em}
\subsection{Semi-Supervised Learning (SSL) for TOD Systems}
\vspace{-0.5em}
There are increasing interests in developing SSL methods for TOD systems, which aims to leverage both labeled and unlabeled data. 
Latent variable modeling is an important SSL approach, which, when applied to semi-supervised TOD systems, has been proposed in \cite{zhang-2020-labes} and developed in \cite{liu2022mga,liu2023variational}. The dialog states and the system acts are treated as latent variables in unlabeled data and the variational algorithm is used for model training. Recently, it is shown that JSA learning significantly outperforms variational learning for semi-supervised performances \cite{cai2022advancing}. Our work is different from previous works in that labels of relevant knowledge in KRTOD systems are modeled as latent variables.

\vspace{-1.0em}
\section{Method}

\subsection{Knowledge-Retrivel TOD Model (KRTOD)} \label{sec:model}

Assume we have a dialog 
with $T$ turns of user utterances and system responses, denoted by $u_{1},r_{1},\cdots,u_{T},r_{T}$ respectively. At turn $t$, based on dialog context, the system queries a task-related knowledge base (KB) to get relevant knowledge, decides its action, and generates appropriate responses.
The KB is composed of entities with attributes, or say, slot-value pairs, denoted by $\{\text{sv}^{1},\text{sv}^{2},\cdots,\text{sv}^{N}\}$.


The slot-value pairs that are relevant for the system to respond at turn $t$ are denoted by $\xi_t$. In labeled data, $\xi_t$ is observed, while in unlabeled dialogs, it becomes a latent variable. 
The system action $a_t$ could be modeled in the same way.
So we denote the latent variables at turn $t$ collectively by $h_{t}= \{\xi_{t} , a_{t}\}$. 
Similar to \cite{cai2022advancing}, a latent state TOD model 
can be defined as follows, with model parameter $\theta$:
\vspace{-1.0em}
\begin{align}
\vspace{-0.5em}
&p_\theta(h_{1:T}, r_{1:T}|u_{1:T}) 
=\prod_{t=1}^T p_\theta(h_{t},r_{t}|c_{t},u_{t})
\label{eq:LS-1}
\end{align}
where $c_{t}=u_{1},r_{1},\cdots,u_{t-1},r_{t-1}$ denotes the dialog context at turn $t$.
The joint model of $h_{t},r_{t}$ is further decomposed into a knowledge retriever $p_\theta^{\text{ret}}$ and a response generator $p_\theta^{\text{gen}}$.
\begin{equation}
p_\theta(h_{t},r_{t}|c_{t},u_{t})
= p_\theta^{\text{ret}}(\xi_t|c_{t}, u_{t}) \times p_{\theta}^{\text{gen}}(a_t,r_t|c_{t}, u_{t}, \xi_{t})
\label{eq:rag}
\end{equation}
From Eq. \ref{eq:rag}, we can clearly see the differences between KRTOD and traditional TOD systems. First, a retrieval model $p_\theta^{\text{ret}}$ is introduced in KRTOD to do knowledge selection procedure, instead of doing traditional database query. Second, traditional dialog state tracking is avoided as shown in Eq. \ref{eq:rag}.
\begin{figure}[t]
\centering
	\includegraphics[width=1\linewidth]{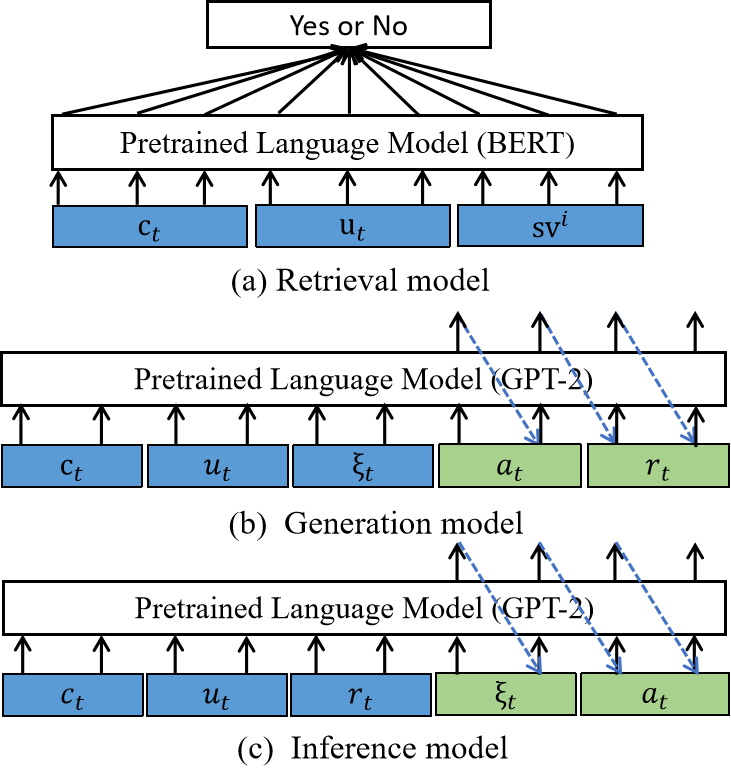}
 	\vspace{-0.5em}
	\caption{The model implementations: (a) the retrieval model, (b) the generation model, and (c) the inference model for our \modelname. All of the variables ${c_t},{u_t},{\xi_t},{a_t},{r_t},{\text{sv}^i}$ are represented by token sequences in our experiments.}
 	\vspace{-0.5em}
	\label{fig:models}
\end{figure}




In order to perform unsupervised learning over unlabeled dialogs (to be detailed below), we introduce an inference model $q_\phi(h_{1:T}|u_{1:T},r_{1:T})$ as follows to approximate the true posterior $p_\theta(h_{1:T}|u_{1:T},r_{1:T})$:
\vspace{-1.0em}
\begin{align}
q_\phi(h_{1:T}|u_{1:T},r_{1:T})=&\prod_{t=1}^{T} q_\phi(h_{t}|c_{t},u_{t},r_{t}) \nonumber \\
=&\prod_{t=1}^T q_\phi(\xi_{t},a_{t}|c_{t},u_{t},r_{t})
\label{eq:posterior}
\end{align}
\vspace{-1.0em}


\vspace{-1.0em}
\subsection{Model Implementation}
\vspace{-0.5em}
To implement the models introduced in Section \ref{sec:model}, we use a BERT \cite{devlin2019bert} based classification model to build the knowledge retriever $p_\theta^{\text{ret}}$ and GPT2 based autoregressive generative models to build both the generation model $p_{\theta}^{\text{gen}}$ and the inference model $q_\phi$, as illustrated in Figure \ref{fig:models}.

The retrieval model $p_\theta^{\text{ret}}$ aims to retrieve relevant knowledge from the KB in order to respond to user utterances.
Particularly, the knowledge piece $\xi_t$ necessary for turn $t$ is represented by $\xi_{t} \triangleq \xi_{t}^{1} \oplus \xi_{t}^{2} \oplus \cdots \oplus \xi_{t}^{N}$, where $\oplus$ denotes sequence concatenation. 
$\xi_{t}^{i} = \text{sv}^{i}$ if the slot-value $\text{sv}^{i}$ is relevant to the response $r_t$; otherwise, $\xi_{t}^{i}$ is set to be empty. 
Moreover, we assume that whether $\text{sv}^{i}$ is contained in $\xi_{t}$ or not is independent of each other. Therefore, the retrieval probability can be written as follows:
\begin{align}
\vspace{-0.5em}
p_\theta^{\text{ret}}(\xi_t|c_{t}, u_{t}) = \prod \limits_{i=1}^{N} p_\theta^{\text{ret}}(\xi_t^{i}|c_{t}, u_{t})
\label{eq:retrieval}
\vspace{-0.5em}
\end{align}
where $p_\theta^{\text{ret}}(\xi_t^{i}|c_{t}, u_{t})$ is realized based on BERT, using $c_{t}\oplus u_{t} \oplus \text{sv}^i$ as input, as shown in Figure \ref{fig:models}. Specifically, denote the mean-pooling of the BERT encoder output by 
\vspace{-0.5em}
\[\mathbf{x} = \text{meanpooling}(\text{BERT}(c_{t}\oplus u_{t} \oplus \text{sv}^i))\]
\vspace{-0.5em}
Then we have
\begin{align*}
&p_\theta^{\text{ret}}(\xi_t^{i} = \text{sv}^i |c_{t}, u_{t}) = \frac{1}{1+e^{-(\mathbf{W} \mathbf{x}+b)}} \\
&p_\theta^{\text{ret}}(\xi_t^{i} = \texttt{NULL}  |c_{t}, u_{t}) = 1- p_\theta^{\text{ret}}(\xi_t^{i} = \text{sv}^i |c_{t}, u_{t})
\end{align*}
where $\mathbf{W},b$ are trainable parameters.

As shown in Figure \ref{fig:models}, both the generation model and the inference model are realized based on GPT2, and the probabilities are calculated in an autoregressive way as follows:
\begin{align}
&{p}_{\theta}^{\text{gen}}(a_{t}, r_{t} \mid c_{t}, u_{t}, \xi_{t}) \nonumber\\
&=\prod_{l=1}^{|a_{t}\oplus r_{t}|}
p_{\theta}^{\text{gen}}(y^l \mid c_{t}, u_{t}, \xi_{t}, y^1, \ldots, y^{l-1}) \label{eq:gpt2_a}\\
&q_{\phi}(\xi_{t}, a_{t} \mid c_{t}, u_{t}, r_{t}) \nonumber\\
&=\prod_{l=1}^{|\xi_{t}\oplus a_{t}|}  q_{\phi}(y^l \mid c_{t}, u_{t}, r_{t} , y^1, \ldots, y^{l-1})
\label{eq:gpt2}
\end{align}
where $|\cdot|$ denotes the length in tokens, and $y^l$ the $l$-th token.

In supervised training, we use the ground truth $\xi_t$, which is annotated in the dataset, to maximize the log probabilities in Eq. \ref{eq:retrieval} - \ref{eq:gpt2}. 
In testing, according to Eq. \ref{eq:rag},  we firstly retrieve relevant slot-value pairs $\xi_t$ by thresholding $p_\theta^{\text{ret}}(\xi_t^{i} = \text{sv}^i |c_{t}, u_{t}), i=1,\cdots,N$; then, we generate $a_t$ and $r_t$, based on retrieved $\xi_t$.
\vspace{-0.5em}
\subsection{Semi-Supervised Training of KRTOD via JSA}
\vspace{-0.5em}
Consider an unlabeled dialog of $T$ turns, where the user utterances and system responses $u_{1:T},r_{1:T}$ are given, but there are no labels for the KB results and system acts, thus being treated as latent variables $h_{1:T} \triangleq \{\xi_{1:T},a_{1:T}\}$. We can use JSA learning to maximize marginal likelihood $p_\theta(r_{1:T}|u_{1:T})$, which iterates Monte Carlo sampling and parameter updating. 
Particularly, we use a recursive turn-level MIS sampler to sample $h_{1:T}$, as developed in \cite{cai2022advancing}. At each turn $t$, the MIS sampler works in a propose, accept or reject way, as follows:

1) Propose $h'_{t} \sim q_\phi(h_{t}|c_{t},u_{t},r_{t})$. 

2) Simulate $\eta \sim \text{Uniform}[0,1]$ and let
\begin{equation} \label{eq:tod-accept-reject}
h_{t} = \left\{
\begin{split}
&h'_{t}, &&\text{if}~\eta \le min\left\lbrace 1, \frac{w(h'_{t})}{w(\bar{h}_{t})} \right\rbrace \\ 
&\bar{h}_t, &&\text{otherwise}
\end{split}
\right.
\end{equation}
where $\bar{h}_t$ denotes the cached latent state, and the importance ratio $w(h_{t})$ between the target and the proposal distribution can be written as:
\begin{align}
w(h_{t}) 
\propto & \frac{p_\theta(h_t,r_t|c_{t}, u_t)}{q_\phi(h_{t}|c_{t},u_t,r_t)} \nonumber \\
=&\frac{p_\theta^{\text{ret}}(\xi_t|c_{t}, u_{t}) \times p_{\theta}^{\text{gen}}(a_t,r_t|c_{t}, u_{t}, \xi_{t}) }{q_\phi(\xi_{t},a_{t}|c_{t},u_t,r_t)} \label{eq:tod-MIS-weight}
\end{align}

Now that we have introduced the method of how to deal with unlabeled data in JSA learning.
Semi-supervised learning over a mix of labeled and unlabeled data could be readily realized in \modelname{} by maximizing the sum of $\log p_\theta(h_{1:T}, r_{1:T}|u_{1:T})$ (the conditional joint log-likelihood) over labeled data and $\log p_\theta(r_{1:T}|u_{1:T})$ (the conditional marginal log-likelihood) over unlabeled data.

The semi-supervised training procedure of \modelname{} is summarized in Algorithm \ref{alg:training}.
Specifically, we first conduct supervised pre-training of both the generative model $p_\theta^{\text{gen}}$ and the inference model $q_\phi$ on labeled data. 


\begin{algorithm}[tb]
  \caption{Semi-supervised training in \modelname{}}
  \label{alg:training}
  \begin{algorithmic}[1]
        \REQUIRE A mix of labeled and unlabeled dialogs.
	    \STATE Run supervised pre-training of $\theta$ and $\phi$ on labeled dialogs;
	    \REPEAT
	    \STATE Draw a dialog $(u_{1:T},r_{1:T})$;
		\IF  {$(u_{1:T},r_{1:T})$ is not labeled}
		\STATE Generate $h_{1:T}$ using the recursive turn-level MIS sampler 
		\ENDIF
		\STATE $J_{\theta}=0, J_\phi=0$; 
        \FOR {$i = 1,\cdots,T$}
        \STATE $J_{\theta} += \log p_{\theta}^{\text{gen}}(a_{t}, r_{t} \mid c_{t}, u_{t}, \xi_{t})$; \label{alg:line:p_gen}
        \STATE $J_\phi += \log q_\phi(\xi_{t}, a_{t} \mid c_{t}, u_{t}, r_{t})$;
        \ENDFOR
        \STATE Update $\theta$ by ascending: $\nabla_\theta J_\theta$;
        \STATE Update $\phi$ by ascending: $\nabla_\phi J_\phi$;
        \UNTIL {convergence}
        \RETURN {$\theta$ and $\phi$}
  \end{algorithmic}
  \vspace{-0.4em}
\end{algorithm}
Then we randomly draw supervised and unsupervised mini-batches from labeled and unlabeled data. For labeled dialogs, the latent states $h_t \triangleq \{\xi_{t},a_t\}$ are given. For unlabeled dialogs, we apply the recursive turn-level MIS sampler to sample the latent states $h_t$ and treat them as if being given.
The gradients calculation and parameter updating are then the same for labeled and unlabeled dialogs. \footnote{Only the response generator $p_\theta^{\text{gen}}$ is updated
in Line \ref{alg:line:p_gen} in Algorithm \ref{alg:training}. After supervised pre-training, the retrieval model $p_\theta^{\text{ret}}$ is not updated over unlabeled data, since the KB is not available for unlabeled data.}
\section{Experiments}
\vspace{-0.5em}
\subsection{Experiment Settings and Baselines}
\vspace{-0.5em}
Experiments are conducted on a real-life human-human dialog dataset, called MobileCS, released from the EMNLP 2022 SereTOD Challenge \cite{ou2022achallenge}. 
The MobileCS dataset is 
from customer-service logs, instead of collected by the Wizard-of-Oz method \cite{budzianowski2018large}. Therefore, building TOD systems over such dataset is more challenging, as user utterances are more casual and may
contain noises from automatic speech recognition \cite{ou2022achallenge}. 
MobileCS contains a total of around 100K dialogs.
The labeled part was officially randomly split into train, development, and test sets, which consist of  8,953, 1014 and 955 dialog samples, respectively. The remaining 87,933 dialogs are unlabeled.
Therefore, experiments in both labeled-only and semi-supervised settings (over both labeled and unlabled data) can be conducted and fairly compared. 
For evaluation, we follow the original scripts in \cite{liu2022information} and mainly focus on two metrics, \emph{Success rate} and \emph{BLEU}, which mainly evaluate the quality of the generated responses. 
\emph{Success rate} measures how often the system is able to provide all the entities and values requested by the user, which is crucial in performing a successful dialog. \emph{BLEU} is used to measure the fluency of the generated responses by analyzing the amount of n-gram overlap between the real responses and the generations. 
The overall performance is measured by \emph{Combined score}, which is \emph{Success} + 2*\emph{BLEU}, as in the original scripts. Several strong baselines are compared with our method. The official baseline \cite{liu2022information} uses predicted dialog state to query the database, which is referred to as KB-query in Table \ref{tab:main-results}. PRIS \cite{zeng2022semi} concatenates the whole local KB to the dialog history, which is referred to as KB-grounded in Table \ref{tab:main-results}. TJU-LMC \cite{yang2022discovering} uses coarse-to-fine intent detection to improve performance over the baseline, while Passion \cite{lu2022passion} improves prompting scheme to boost performance. PRIS, Passion, and TJU-LMC were top three teams in the SereTOD Challenge.
In our experiments, to follow the Challenge guideline, hyper-parameters are chosen based on the development set, and evaluated on the test set.

\vspace{-1.0em}
\subsection{Main Results}
\vspace{-0.5em}
As shown in Table \ref{tab:main-results}, our method achieves SOTA results on MobileCS in both labeled-only and semi-supervised settings. The results clearly show the superiority of our method, which greatly outperforms the baseline method (KB-query) and the competitive KB-grounded method, especially in the Success rate metric. It is evident that importing a knowledge retriever into a TOD system can significantly enhance the system's ability in accurately providing the necessary knowledge for the current turn, which is the key to achieve high Success rate.

From the comparison between the semi-supervised and the labeled-only results in Table \ref{tab:main-results}, we can see that performing semi-supervision using unlabeled data is crucial for improving the system's performance. Using the proposed \modelname{} to perform semi-supervision can bring substantial gain in both the Success rate and BLEU-4 metrics, and improves over the strong KRTOD baseline by \textbf{7.07} in the Combined score.

It can be seen from Table \ref{tab:main-results} that although our model achieves the SOTA results on Combined score and Success rate, the BLEU-4 score is not so competitive. Presumably, the main reason for the difference in the BLEU-4 performance lies in the number of model parameters. Larger model (T5 1B) can better fit the dataset, resulting higher BLEU score. The comparison between the KB-grounded method and our proposed method using the same model (GPT2 100M) in the Labeled-only part of Table \ref{tab:main-results} shows that the two methods have similar BLEU scores, while our method improves greatly on the Success rate. Also note that PRIS \cite{zeng2022semi} uses large-scale customer-service dialogs to pre-train their model (T5 1B), which further improves the BLEU score over their GPT2 based model. Therefore, further work can be done to combine our proposed method with a larger backbone.
\begin{table}[t]
    \caption{Main results on the MobileCS dataset. Success, BLEU-4, and Combined score are reported. Our approach achieves SOTA results on both labeled-only and semi-supervised settings. Within the parentheses show the backbone models and their number of parameter.}
    \vspace{-1.0em}
	\centering
	\resizebox{1\linewidth}{!}{
			\begin{tabular}{c c c c c}
				\toprule
				Setting &Method  &Success &BLEU-4 &Combined \\ 
				\midrule
  \multirow{5}{*} {Semi-supervised} & Baseline \cite{liu2022information}  & 31.5  &4.170  &39.84\\
   &Passion \cite{lu2022passion} &43.2   &6.790  &56.78\\
   &TJU-LMC \cite{yang2022discovering}& 68.9 &7.54  &83.98\\
&PRIS \cite{zeng2022semi} & 78.9 &14.51  &107.92\\
&\modelname{} & 91.8 & 9.677  &\textbf{111.15}\\
\midrule
\midrule
 \multirow{4}{*} {Labeled-only} &KB-query (GPT2 100M) \cite{liu2022information} & 31.5  &4.170  &39.84\\
 &KB-grounded (GPT2 100M) \cite{zeng2022semi} & 64.2  &8.845  &81.89\\
&KB-grounded (T5 1B) \cite{zeng2022semi} & 74.1  &11.32  &96.74\\
&KRTOD (GPT2 100M)  &86.8   &8.639  &\textbf{104.08}\\
				\bottomrule
    \vspace{-0.5em}
		\end{tabular}}
	\vspace{-1.0em}
		\label{tab:main-results}
\end{table}
\vspace{-0.5em}
\begin{table}[t] \tiny- 
    \caption{Comparison between pseudo labeling (PL) and JSA learning methods. Ratio means the ratio between the number of unlabeled dialogs and the number of labeled dialogs in training. p-value means the significant test result for Combined score.} 
        \vspace{-1.0em}
	\centering
	\resizebox{1\linewidth}{!}{
			\begin{tabular}{cc cccc cccc}
				\toprule
				Ratio &Method  &Success &BLEU-4 &Combined &p-value \\ 
				\midrule
				\multirow{2}{*}{1:1} 
				&\comparemodel{}    & 87.5  &  8.853 & 105.21 &\multirow{2}{*}{0.589}\\
				& JSA  &  88.0 & 8.713 & \textbf{105.43} \\
				\midrule
				\multirow{2}{*}{2:1} 
				&\comparemodel{}  & 87.8  & 9.196 & 106.19 &\multirow{2}{*}{0.853}\\
				&JSA   & 88.7 & 9.490&   \textbf{107.68} \\
				\midrule
				\multirow{2}{*}{4:1} 
				&\comparemodel{}       & 88.5  & 9.341  & 107.18 &\multirow{2}{*}{0.037}\\
				&JSA   &90.9 & 9.398 &\textbf{109.70}\\
				\midrule
				\multirow{2}{*}{9:1} 
				&\comparemodel{}     & 89.4   & 9.532 & 108.46 &\multirow{2}{*}{0.055}\\
				&JSA  & 91.8 &9.677& \textbf{111.15}
\\
				\bottomrule
    \vspace{-2.5em}
		\end{tabular}}
	\vspace{-2.0em}
		\label{tab:compare-results}
\end{table}

\vspace{-1.5em}
\subsection{Analysis and Ablation}
\vspace{-0.5em}
We further examine whether our semi-supervised method \modelname{} is competitive among other semi-supervised methods. We mainly compare our method with a classic semi-supervised method, called pseudo labeling (PL) or self-training (ST), which has been used in \cite{zeng2022semi} and \cite{liu2023variational} for TOD systems.
We implement the PL method, similar to the ``ST with inference model'' method in \cite{liu2023variational}.
The results in Table \ref{tab:compare-results} show that \modelname{} outperforms PL constantly in all ratios. 
The relative improvement of JSA over PL in reducing errors in Success rate is 23\% under ratio 9:1.
Further, the p-values from the matched-pairs significance tests \cite{gillick1989some} in Combined score show that as the size of unlabeled data increases, the improvements of \modelname{} over PL become more significant, confirming the superiority of \modelname{} in leveraging unlabeled data.
\vspace{-0.5em}
\section{Conclusion and Future Work}
\vspace{-0.5em}
In this paper, we propose KRTOD, which introduces a knowledge retriever into the TOD system. 
The knowledge retriever relieves the burden to track the dialog state and the inflexibility of using the dialog state to query the database in the traditional TOD systems. 
The proposed KRTOD method  greatly outperforms previous strong baseline methods over the challenging real-life MobileCS dataset.
Moreover, by combining KRTOD with JSA based semi-supervised learning, our semi-supervised TOD system, \modelname{}, improves over the strong supervised baseline and achieves SOTA results on MobileCS. 
For future work, KRTOD potentially can exploit more types of knowledge sources, such as passages, documents and knowledge graphs, in addition to slot-value pairs used in this paper.


\bibliographystyle{IEEEtran}
\bibliography{mybib}

\clearpage
\appendix

\end{document}